\documentclass[runningheads]{llncs}
\usepackage[T1]{fontenc}
\usepackage{booktabs}  

\usepackage{makecell} 
\usepackage{amsmath}  
\usepackage{amssymb}  
\usepackage{hyperref}

\usepackage{graphicx,verbatim}
\usepackage{color}

\begin{document}
\title{Flip Distribution Alignment VAE for Multi-Phase MRI Synthesis}
%

\author{
Xiaoyan Kui\inst{1} \and
Qianmu Xiao\inst{1}\thanks{Corresponding author.} \and
Qinsong Li\inst{2} \and
Zexin Ji\inst{1} \and
Jielin Zhang\inst{3} \and
Beiji Zou\inst{1}
}
\authorrunning{Kui et al.}

\institute{
School of Computer Science and Engineering, Central South University, ChangSha, China \\
\email{\{xykui, qianmu, zexin.ji, bjzou\}@csu.edu.cn} \and
Big Data Institute, Central South University, ChangSha, China \\
\email{qinsli.cg@foxmail.com} \and
Department of Radiology, The Second Xiangya Hospital, Central South University, ChangSha, China \\
\email{238212362@csu.edu.cn}
}

\maketitle              
\begin{abstract}
Separating shared and independent features is crucial for multi-phase contrast-enhanced (CE) MRI synthesis. 
However, existing methods use deep autoencoder generators with low parameter efficiency and lack interpretable training strategies. 
In this paper, we propose Flip Distribution Alignment Variational Autoencoder (FDA-VAE), 
a lightweight feature-decoupled VAE model for multi-phase CE MRI synthesis. 
Our method encodes input and target images into two latent distributions that are symmetric concerning a standard normal distribution, 
effectively separating shared and independent features. 
The Y-shaped bidirectional training strategy further enhances the interpretability of feature separation.
Experimental results show that compared to existing deep autoencoder-based end-to-end synthesis methods, 
FDA-VAE significantly reduces model parameters and inference time while effectively improving synthesis quality. 
The source code is publicly available at \url{https://github.com/QianMuXiao/FDA-VAE}.

\end{abstract}

\section{Introduction}

Multi-phase contrast-enhanced (CE) MRI provides essential diagnostic information for assessing organ lesions, tumors, and vascular abnormalities. 
However, this imaging technique is still limited by the long scanning time and nephrotoxicity risk caused by gadolinium-based contrast agents.
Medical image super-resolution~\cite{deform,borges2024using} and synthesis~\cite{chu2024anatomic,phan2024structural} techniques address these challenges by generating high-quality or missing images from low-quality or existing scans. 
In CE MRI, synthesizing different enhanced-phase images from unenhanced-phase can effectively reduce scanning time and mitigate the health risks associated with contrast agents.

Current medical image synthesis methods can be categorized into Diffusion-based~\cite{brain-diffusion,monai2024latentdiffusion,common-uni} and Autoencoder-based (AE-based) approaches~\cite{pix2pix,resvit,ptnet,hi-net,i2i-mamba}. 
Diffusion-based methods generate high-quality synthetic images but require substantial computational resources and are constrained by inflexible training strategies. 
In contrast, AE-based methods provide greater flexibility in training by allowing direct manipulation of latent space structures and optimization objectives.

\begin{figure}[t]
    \includegraphics[width=\textwidth]{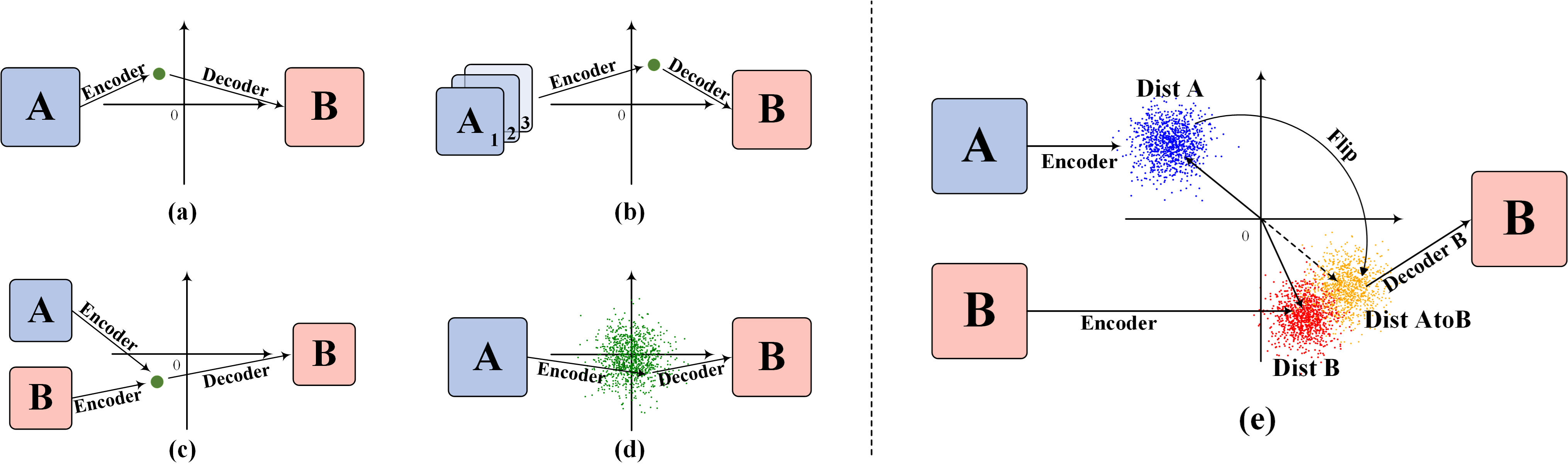}
    \caption{Overview of existing AE-based medical image synthesis strategies. 
        \textbf{(a)} Basic one-to-one autoencoder. 
        \textbf{(b)} Multi-phase many-to-one autoencoder. 
        \textbf{(c)} Autoencoder with latent space comparative learning. 
        \textbf{(d)} Variational Autoencoder (VAE). 
        \textbf{(e)} Our proposed method: Flip Distribution Alignment Variational Autoencoder (FDA-VAE).} 
        \label{fig1}
\end{figure}

Existing AE-based models employ different feature extractors to improve representation learning. 
Traditional CNN-based AEs~\cite{pix2pix} (Fig.\ref{fig1} (a)) suffer from limited receptive fields, 
making it difficult to capture long-range dependencies. 
Vision Transformers (ViTs)~\cite{vit,ptnet} address this issue but introduce high computational costs. 
Recently, state-space models (SSMs) such as Mamba~\cite{mamba,i2i-mamba} have emerged as efficient alternatives. 
Some studies further integrate hybrid architectures~\cite{resvit,hi-net,mt-net,mustgan,transunet} that combine CNNs and Transformers to balance efficiency and performance. 
However, optimizing the encoding-decoding structure remains crucial for high-quality synthesis, as shown in Fig.\ref{fig1} (b)-(d).  
Some approaches integrate multi-phase input features to generate the target modality Fig.\ref{fig1} (b). 
Others use latent feature contrastive learning~\cite{park2020contrastive} Fig.\ref{fig1} (c) or structure-supervised loss~\cite{mask-aware,mask-gan} to reinforce shared features between the input and target modalities. 
Furthermore, some studies encode latent features as probabilistic distributions~\cite{vae,cackowski2023imunity,laptev2021generative,cetin2023attri} Fig.\ref{fig1} (d). 
This improves the smoothness of the latent space. It also enhances uncertainty modeling and generation diversity.  

However, there are still some problems with these strategies:
(1) Many-to-one methods require paired multi-phase data for training. 
(2) One-to-one methods focus only on cross-modality mappings or shared features but neglect their independent features. 
(3) Existing methods still rely on deep autoencoders for cross-modality mapping even under limited paired training data, which may lead to suboptimal parameter utilization.

To address these problems, we propose Flip Distribution Alignment VAE (FDA-VAE).
It is a lightweight feature-decoupled model for multi-phase CE MRI synthesis.
Our method uses a compact hybrid-architecture VAE as the generator to reduce model parameters and improve efficiency.
We introduce Flip Distribution Alignment (FDA) as a structured constraint on the latent space.
Specifically,  our method encodes input and target images as two latent distributions, 
enforcing symmetry by setting opposite means and equal variances.
This ensures that shared features are preserved while maximizing independent components, 
with transformation achieved via simple mean flipping.
Additionally, we design a Y-shaped bidirectional training strategy, 
enabling both self-reconstruction and cross-phase synthesis through mean flipping.
This enhances the interpretability and stability of latent space modeling.
Compared to existing methods, 
FDA-VAE provides a structured and interpretable latent space representation, 
significantly improving synthesis quality and parameter efficiency.

\section{Method}\label{sec2}

\begin{figure}[t]
    \includegraphics[width=\textwidth]{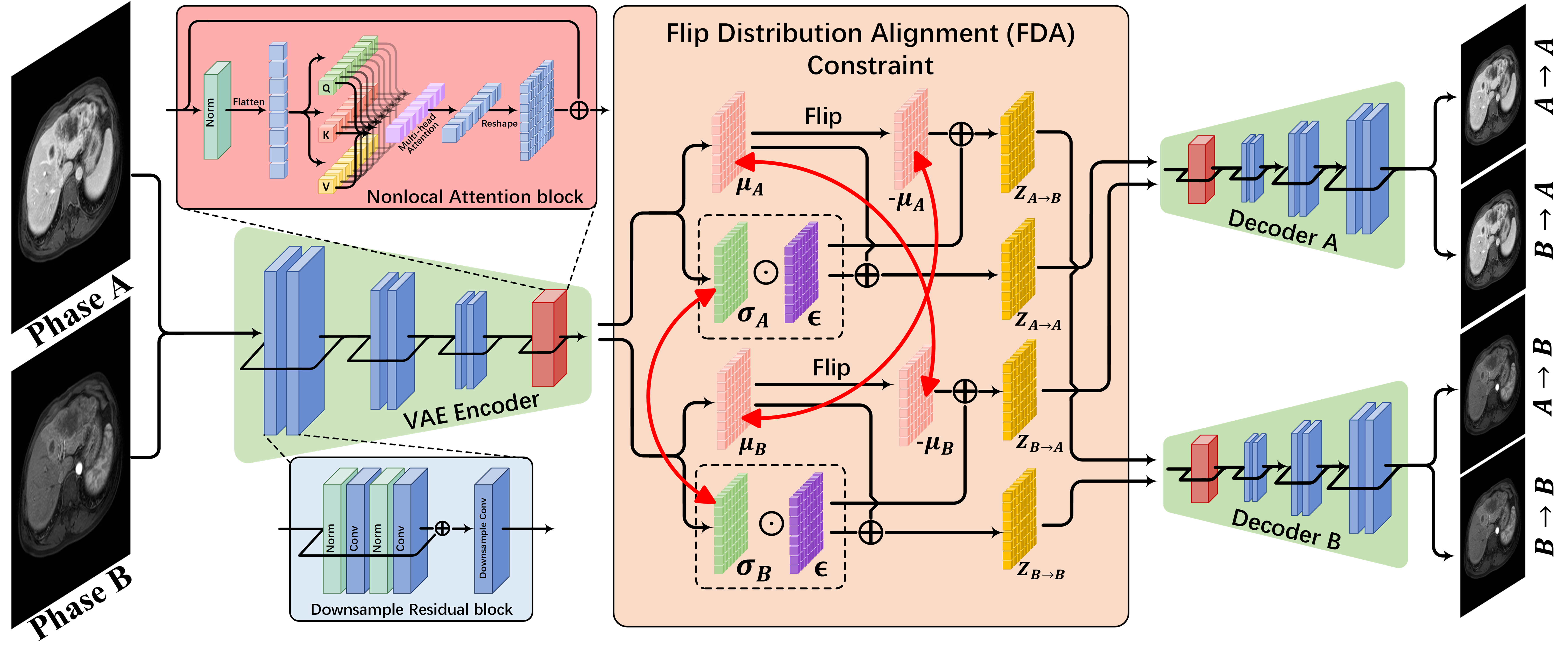}
    \caption{
        Overview of the proposed Flip Distribution Alignment Variational Autoencoder (FDA-VAE). 
        The model consists of a shared encoder, two independent decoders, 
        and a flip distribution alignment (FDA) constraint layer. 
        During training, a pair of different phase MRI images is input to obtain self-reconstructed and cross-phase transformed outputs. 
        In the inference stage, only the target decoder is retained. 
        The image is encoded to obtain the latent distribution, 
        and the mean vector is flipped before decoding, generating the target-phase image from the flipped distribution.
        } 
    \label{fig2}
\end{figure}
\textbf{Lightweight VAE vs Deep AE.} 
Pre-trained  VAE~\cite{vae} demonstrates excellent data compression and decoding capabilities in high-resolution image synthesis tasks.
Recent approaches, such as Latent Diffusion (LDM)~\cite{latent-diffusion} and Visual Autoregressive~\cite{tian2025visual} (VAR), 
utilize pre-trained VAE or Vector Quantised-VAE (VQ-VAE)~\cite{van2017neural} models. 
These models typically contain around 100M parameters and are trained on 1.2 million natural images. 
They are first trained for image self-reconstruction, 
providing a latent representation that facilitates subsequent feature generation. 
In medical image synthesis, MONAI's LDM-based approach~\cite{monai2024latentdiffusion} uses about 38,000 brain MRI slices to train a VAE with about 12M parameters.
In contrast, methods like ResVit~\cite{resvit} and I2I-Mamba~\cite{i2i-mamba} train cross-modality mappings using only about 2,500 paired slices, yet rely on generators exceeding 100M parameters.
Although the self-reconstruction task is relatively simple, cross-modality medical images typically exhibit strong structural correlations. 
In terms of efficiency, existing deep AE generators tend to have excessive parameters, resulting in inefficient utilization, particularly given the limited size of medical datasets.

In this paper, we propose utilizing a shallow, lightweight VAE backbone directly as the generator, 
aiming to enhance image synthesis quality and model interpretability through structured latent space modeling.
As shown in Fig.~\ref{fig2}, we construct a hybrid-architecture VAE backbone, 
where both the encoder and decoder consist of three residual convolutional blocks and one non-local attention block to capture local and global features. 
Compared to existing deep AE generators, our backbone has fewer layers and a narrower model width. 
The formula for the model is as follows: 
\begin{equation}
    \mu, \sigma = \text{Encoder}(x),\quad z =\mu + \sigma * \epsilon,\quad \epsilon \sim \mathcal{N}(0,1),\quad \hat{x} = \text{Decoder}(z)
    \label{eq1}
\end{equation}
\begin{equation}
    \mathcal{L}_{\text{Kullback-Leibler}}\big(\mathcal{N}(\mu, \sigma^2) \parallel \mathcal{N}(0, 1)\big) = 
    \frac{1}{2} \Big( \mu^2 + \sigma^2 - \log(\sigma^2) - 1 \Big)
    \label{eq2}
\end{equation}
Given an input image \(x\), the encoder outputs a mean vector \(\mu\) and a variance vector \(\sigma\). 
Latent features \(z\) are then sampled from this latent distribution \(\mathcal{N}(\mu, \sigma^2)\), and subsequently decoded into the target image \(\hat{x}\) (Eq.\ref{eq1}).
Additionally, we employ the Kullback--Leibler (KL) divergence constraint (Eq.\ref{eq2}) to regularize the encoded distributions towards a standard normal distribution.\\

\begin{figure}[!h]
    \includegraphics[width=\textwidth]{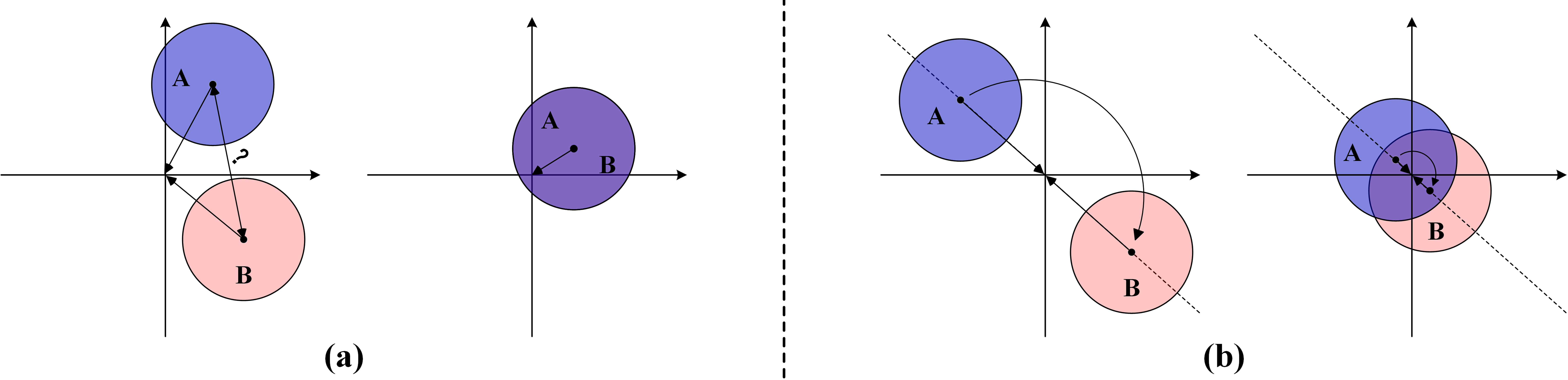}
    \caption{Convergence process of input and target distributions: (a) KL divergence only, (b) KL divergence + FDA.} 
    \label{fig3}
\end{figure}

\noindent
\textbf{Flip Distribution Alignment (FDA).}
The core idea of FDA-VAE is to build a structured and efficient latent space representation. 
With a shared encoder-decoder, we map the input and target images to separate latent distributions and sample from them to synthesize the target image. 
However, solely relying on KL divergence for regularization leads to two issues, as shown in Fig.\ref{fig3} (a). 

In the early stages, the input (A) and target (B) distributions move toward the standard normal distribution independently. 
Without explicit constraints, they approach from random locations, leading to unpredictable relative positioning. 
This misalignment disrupts feature correspondence and increases divergence, 
making feature transformation difficult and degrading synthesis quality. 
As training progresses, the lack of alignment causes the distributions to collapse onto each other, 
overemphasizing shared features while suppressing modality-specific information, 
reducing independent feature distinctiveness.
To address these problems, we introduce an additional Flip Distribution Alignment (FDA) constraint shown in Fig.\ref{fig3} (b). 
FDA constrains the input and target distributions to remain symmetric concerning the standard normal distribution throughout training. 
Specifically, we enforce equal variances ($\sigma_A^{2}=\sigma_B^{2}$) and opposite means ($\boldsymbol{\mu}_A = -\boldsymbol{\mu}_B$) as show in Eq.\ref{eq3}.  
\begin{equation}
    \mathcal{L}_{FDA} = 
    \bigl\|\boldsymbol{\mu}_A + \boldsymbol{\mu}_B\bigr\|_{1} + 
    \bigl\|\sigma_A^{2} - \sigma_B^{2}\bigr\|_{1}
    \label{eq3}
\end{equation}
Combining KL divergence and FDA constraints ensures that both distributions converge toward the standard normal distribution while maintaining structural symmetry. 
This design ensures that the input and target features remain maximally separated during convergence while preserving alignment with the standard normal distribution.
Additionally, the symmetric relative positioning allows feature transformation to be efficiently performed via a simple mean-flipping operation. \\

\noindent
\textbf{Y-shaped Bidirectional Training.} 
To further enhance feature disentanglement, we design a Y-shaped bidirectional training strategy, consisting of a shared encoder that maps both modalities to symmetric latent distributions and two phase-specific decoders to synthesize images.
Given input phase $x_A$, the process involves encoding, flipping, and decoding to obtain $x_{A\to A}$ and $x_{A\to B}$, while input $x_B$ follows the same process for $x_{B \to B}$ and $x_{B \to A}$. 
\begin{equation}
    \mu_A,\sigma_A = \text{Encoder}(x_A),\quad z_{A \to A} \sim \mathcal{N}(\mu_A,\sigma_A^2),\quad z_{A \to B} \sim \mathcal{N}(-\mu_A,\sigma_A^2)
\end{equation}
\begin{equation}
    \hat{x}_{A \to A} = \text{Decoder}_A(z_{A \to A}), \quad \hat{x}_{A \to B} = \text{Decoder}_B(z_{A \to B})
\end{equation}
For the self-reconstruction task, we use L1 loss for supervision $\mathcal{L}_{Rec}$, for the cross-phase synthesis task, we incorporate L1 loss $\mathcal{L}_{Trans}$, GAN loss $\mathcal{L}_{GAN}$, and perceptual loss $\mathcal{L}_{Perce}$ for co-supervision. 
The entire loss function FDA-VAE is summarized as $\mathcal{L}_{FDA-VAE}$:
\begin{equation}
\begin{aligned}   
    \mathcal{L}_{FDA-VAE} = & \lambda_{\text{rec}}\mathcal{L}_{\text{Rec}} + \mathcal{L}_{\text{Tran}} + \lambda_{\text{gan}}\mathcal{L}_{\text{GAN}} 
    \\ & + \lambda_{\text{perce}}\mathcal{L}_{\text{Perce}} + \lambda_{\text{kl}}\mathcal{L}_{\text{KL}} + \lambda_{\text{fda}}\mathcal{L}_{\text{FDA}}
\end{aligned}
\end{equation}
where $\lambda_{rec},~\lambda_{gan},~\lambda_{perce}$ and $\lambda_{fda}$,  take the value of $1\times10^{-2}$, $\lambda_{kl}$ takes the value of $1\times10^{-7}$.

\section{Experiments}
\subsection{Experiment Setups}
\textbf{Dataset \& Pre-process.} 
FDA-VAE was trained on the LLD-MMRI 2023 dataset~\cite{lldmmri}, 
containing 498 patients across seven liver lesion types (four benign, three malignant).
We selected four T1 contrast-enhanced phases: Pre-contrast (Pre), arterial (CA), venous (CV), and delayed (Delay), 
designing six early-to-late phase synthesis tasks.
Non-rigid registration was performed using ANTsPy~\cite{ANTsPy} with the C+V phase as the reference.
To ensure lesion-type consistency, images were grouped by disease category and split 4:1 for training and validation.
Preprocessing included top 0.1\% intensity clipping, normalization, and resizing to 256×256. \\
\textbf{Evaluation Metrics.} 
We evaluated our model using PSNR, SSIM~\cite{ssim} and LPIPS~\cite{lpips} to assess image quality. 
Additionally, we analyzed model efficiency in terms of parameter count and inference time per slice.\\
\textbf{Training Details.} 
All experiments were implemented in PyTorch v2.5.1 in conjunction with the MONAI \cite{monai} framework. 
We employed the Adam optimizer with an initial learning rate of 1e-4 and trained each model for 40 epochs on a Linux workstation with 4 $\times$ NVIDIA RTX 4090 24G GPUs. 
It took about six and a half hours to train FDA-VAE.
\subsection{Ablation Study}
We conduct an ablation study to assess the impact of each component.
First, we establish the lightweight VAE backbone as a baseline (Tab.\ref{table1}, VAE (backbone)).
While it achieves reasonable performance, its synthesis quality is constrained by the reduced model capacity. 
Next, we introduce the FDA constraint to enforce structured latent space alignment (Tab.\ref{table1} VAE (KL+FDA)). 
This variant applies to mean flipping to transform the input distribution while using the target distribution for self-reconstruction. 
As shown in the result, the FDA constrained improves PSNR and SSIM while reducing LPIPS, demonstrating its effectiveness in feature separation and alignment. 
Finally, our complete model, FDA-VAE, integrates a bidirectional synthesis training strategy with two independent decoders. 
This further enhances synthesis quality, achieving the highest PSNR and SSIM while maintaining the lowest LPIPS scores (Tab.\ref{table1}). 
These results confirm the role of bidirectional training in stabilizing latent space modeling and improving synthesis performance.  
\subsection{Comparison with state-of-the-art models.}
\setlength{\fboxsep}{1pt}
\begin{table}[!h]
    \centering
    \caption{Overview of Evaluation Results (\textbf{Bold} indicates optimal, \underline{Underline} indicates sub-optimal, \fbox{Box} indicates optimal among the compared models, same as Table 2.)}
    \label{table1}
    \begin{tabular}{lcccccc} 
    \toprule
    \textbf{Method/Task} & Pre$\to$CA & Pre$\to$CV & Pre$\to$Delay 
                        & CA$\to$CV & CA$\to$Delay & CV$\to$Delay\\
    \midrule
    
    \multicolumn{7}{c}{\textbf{PSNR(dB)}$\uparrow$} \\
    Pix2Pix~\cite{pix2pix}       & 24.79 & 23.74 & 23.60 & 24.73 & 24.56 & 27.27 \\
    ResVit~\cite{resvit}         & \fbox{25.34} & 24.06 & 24.33 & 25.88 & 25.41 & 26.63 \\
    TransUnet~\cite{transunet}   & 25.01 & \fbox{24.74} & \fbox{24.74} & \fbox{26.18} & \fbox{25.86} & 26.35 \\
    PTNet~\cite{ptnet}           & 24.85 & 23.68 & 24.03 & 25.38 & 24.68 & \fbox{27.89} \\
    I2I-Mamba~\cite{i2i-mamba}   & 24.95 & 24.39 & 24.12 & 25.46 & 25.24 & 25.61 \\
    \midrule
    VAE(backbone) & 25.23 & \underline{24.96} & 23.63 & 26.48 & 24.65 & 27.08 \\
    VAE (KL+FDA)       & \underline{25.71} & 24.95 & \textbf{25.07} & \underline{26.54} & \underline{26.10} & \underline{27.99} \\
    \textbf{FDA-VAE(Ours)}  & \textbf{25.89} & \textbf{24.98} & \underline{24.89}  & \textbf{26.72} & \textbf{26.33} & \textbf{28.59} \\
    \toprule
    
    \multicolumn{7}{c}{\textbf{SSIM(\%)}$\uparrow$} \\
    Pix2Pix~\cite{pix2pix}       & 80.41 & 68.23 & 77.14 & 78.50 & 78.95 & 84.56 \\
    ResVit~\cite{resvit}         & \fbox{81.79} & 76.32 & \fbox{78.99} & 79.20 & 81.90 & 84.98 \\
    TransUnet~\cite{transunet}   & 81.22 & \fbox{79.79} & 78.84 & \fbox{82.38} & \fbox{\underline{83.07}} & 84.64 \\
    PTNet~\cite{ptnet}           & 81.35 & 77.56 & 78.69 & 81.75 & 80.08 & \fbox{86.19} \\
    I2I-Mamba~\cite{i2i-mamba}   & 81.16 & 76.91 & 78.51 & 79.44 & 81.47 & 82.85 \\
    \midrule
    VAE(backbone) & 72.72 & 79.60 & 76.29 & 82.57 & 74.23 & 80.42 \\
    VAE (KL+FDA)       & \underline{82.99} & \underline{80.10} & \textbf{81.32} & \underline{83.07} & 82.99 & \underline{86.41} \\
    \textbf{FDA-VAE(Ours)}  & \textbf{83.70} & \textbf{80.68} & \underline{81.17} & \textbf{84.01} & \textbf{83.84} & \textbf{87.48} \\
    \toprule
    
    \multicolumn{7}{c}{\textbf{LPIPS}$\downarrow$} \\
    Pix2Pix~\cite{pix2pix}      & 0.0854 & 0.0903 & 0.0914 & 0.0837 & 0.0851 & 0.0568 \\
    ResVit~\cite{resvit}        & \fbox{\textbf{0.0713}} & 0.0776 & \fbox{0.0774} & \fbox{\underline{0.0626}} & \fbox{0.0682} & 0.0534 \\
    TransUnet~\cite{transunet}  & 0.0768 & 0.0761 & 0.0791 & 0.0637 & 0.0677 & 0.0632 \\
    PTNet~\cite{ptnet}          & 0.0771 & 0.0787 & 0.0824 & 0.0635 & 0.0764 & \fbox{\textbf{0.0463}} \\
    I2I-Mamba~\cite{i2i-mamba}  & 0.0739 & \fbox{0.0750} & 0.0783 & 0.0646 & 0.0698 & 0.0632 \\
    \midrule
    VAE(backbone) & 0.0799 & 0.0779 & 0.1064 & 0.0661 & 0.0818 & 0.0576 \\
    VAE (KL+FDA)       & \underline{0.0720} & \underline{0.0735} & \textbf{0.0716} & 0.0656 & \underline{0.0667} & 0.0490 \\
    \textbf{FDA-VAE(Ours)}  & \textbf{0.0713} & \textbf{0.0729} & \underline{0.0723} & \textbf{0.0611} & \textbf{0.0620} & \underline{0.0465} \\
    \bottomrule
    \end{tabular}
\end{table}
\begin{table}[!h]
    \centering
    \caption{Generator Params \& Inference Times.}
    \label{table2}
    \begin{tabular}{lcccccc}
    \toprule
            & Pix2Pix & ResVit & TransUnet & PTNet & I2I-Mamba & \textbf{Ours} \\
    \midrule
    \textbf{Params} (m) & 51.89 & 117.72 & 100.45 & \underline{26.83} & 100.64 & \textbf{11.78} \\
    \textbf{Inference} (secs/slice) & \textbf{0.0019} & 0.0121 & 0.0109 & 0.0141 & 0.0071 & \underline{0.0050} \\
    \bottomrule
    \end{tabular}
\end{table}
\begin{figure}[!h]
    \includegraphics[width=\textwidth]{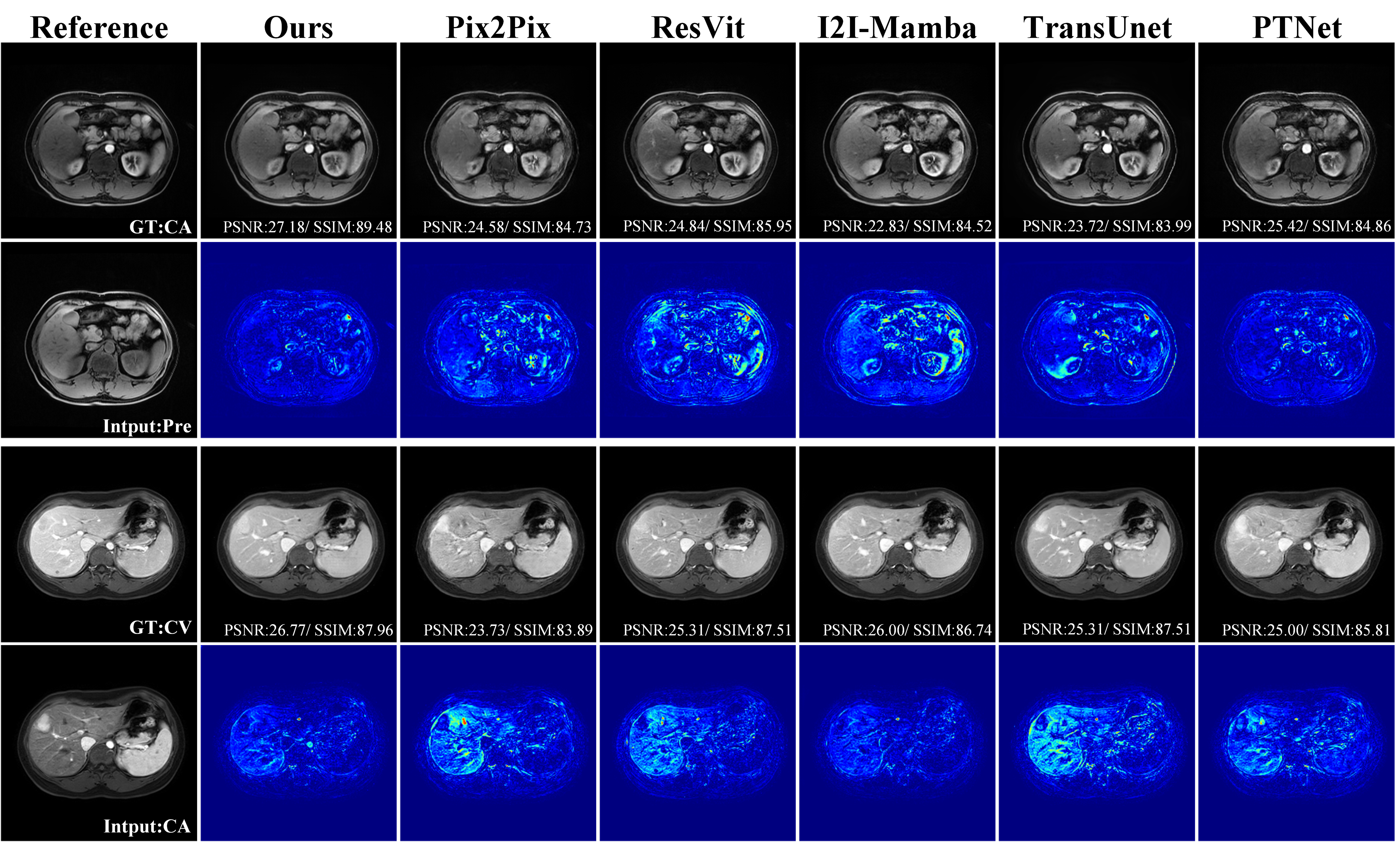}
    \caption{
        Visualization of synthesis result and errors heat maps. 
        } 
    \label{fig4}
\end{figure}
\begin{figure}[!h]
    \includegraphics[width=\textwidth]{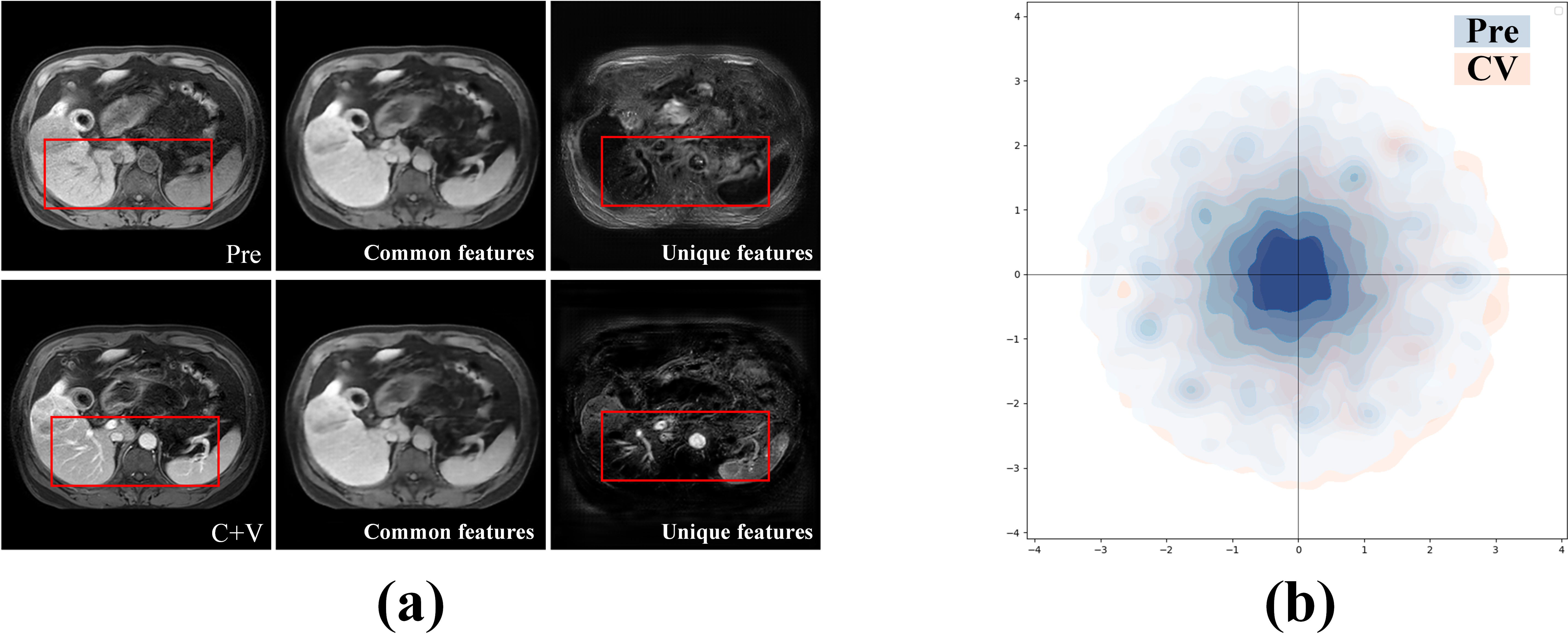}
    \caption{FDA feature decoupling in pixel level (a) \& latent space level (b).} 
    \label{fig5}
\end{figure}
\textbf{Quantitative Analysis.} 
We evaluate the synthesis performance of FDA-VAE against Pix2Pix~\cite{pix2pix}, ResVit~\cite{resvit}, TransUnet~\cite{transunet}, PTNet~\cite{ptnet}, and I2I-Mamba~\cite{i2i-mamba} across six tasks.
Tab.\ref{table1} presents the PSNR, SSIM, and LPIPS results, while Tab.\ref{table2} compares parameter size and inference time.
Among existing methods, ResVit~\cite{resvit} and TransUnet~\cite{transunet} achieve the best synthesis quality but require over 100M parameters and have inference times exceeding 0.01s per slice.
In contrast, our lightweight VAE backbone, with only 11.78M parameters, achieves comparable synthesis quality.
Further improvements are observed with FDA and Y-shaped bidirectional training, significantly enhancing evaluation metrics.
FDA-VAE achieves the best overall performance across most tasks, demonstrating its effectiveness in balancing synthesis quality and computational efficiency. \\
\textbf{Qualitative Analysis.} 
Fig.\ref{fig4} presents the visualization results and error heat maps for six early-to-late phase synthesis tasks across all methods.
Our method achieves the lowest pixel-level error, as indicated by the error heat maps. 
Fig.\ref{fig5} illustrates feature decoupling at both the pixel level (\textbf{a}) and latent space level (\textbf{b}). 
At the pixel level, our method effectively captures common structural and contrast features (second column in \textbf{a}) while preserving phase-specific contrast details (third column in \textbf{a}). 
In latent space, dimensionality reduction visualization shows overlapping regions between input and target distributions, 
with preserved non-overlapping areas, aligning with our FDA design objective. 
\section{Conclusion}
In this paper, we propose a lightweight feature-decoupled VAE framework called FDA-VAE for multi-phase MRI synthesis. 
By incorporating the FDA constraint and a Y-shaped bidirectional training strategy, FDA-VAE simultaneously retains both common and independent features of the input and target images at the latent feature level. 
Compared with state-of-the-art methods, FDA-VAE achieves better synthesis quality while significantly reducing model parameters and inference time. 
In future work, we aim to extend this framework to unpaired data by leveraging unsupervised learning techniques for cross-phase synthesis.

    



\begin{credits}

\subsubsection{\ackname}
This work is supported by the National Natural Science Foundation of China (No. U22A2034, 62177047, 62302530), 
High Caliber Foreign Experts Introduction Plan funded by MOST, 
Key Research and Development Programs of Department of Science and Technology of Hunan Province (No. 2024JK2135), 
Major Program from Xiangjiang Laboratory (No. 23XJ02005), 
the Scientific Research Fund of Hunan Provincial Education Department (No. 24A0018), 
Hunan Provincial Natural Science Foundation (No. 2023JJ40769), 
and Central South University Research Programme of Advanced Interdisciplinary Studies (No. 2023QYJC020).

\subsubsection{\discintname}
The authors have no competing interests to declare that are relevant to the content of this article.

\end{credits}

%
%
%
\bibliographystyle{splncs04}
\bibliography{Paper-1536}
\end{document}